\documentclass[conference]{IEEEtran}
\IEEEoverridecommandlockouts
\usepackage{cite}
\usepackage{amsmath,amssymb,amsfonts}
\usepackage{algorithmic}
\usepackage{graphicx}
\usepackage{textcomp}
\usepackage{xcolor}
\usepackage{subcaption}
\usepackage{booktabs}
\usepackage{hyperref}
\usepackage[headsepline]{scrlayer-scrpage} 
\clearpairofpagestyles 

\def\BibTeX{{\rm B\kern-.05em{\sc i\kern-.025em b}\kern-.08em
    T\kern-.1667em\lower.7ex\hbox{E}\kern-.125emX}}

\begin{document}

\title{Towards Generalizable Deepfake Image Detection with Vision Transformers\\
\thanks{This work has been submitted to the IEEE for possible publication. Copyright may be transferred without notice, after which this version may no longer be accessible.}
}

\author{\IEEEauthorblockN{Kaliki V. Srinanda$^\dagger$, M Manvith Prabhu$^\dagger$, Hemanth K. Mogilipalem$^\S$, Jayavarapu S. Abhinai$^*$, \\Vaibhav Santhosh$^*$, Aryan Herur$^\dagger$, Deepu Vijayasenan$^\dagger$}
\IEEEauthorblockA{\textit{Department of Electronics and Communication Engineering$^\dagger$} \\
\textit{Department of Information Technology$^\S$}\\
\textit{Department of Electrical and Electronics Engineering$^*$}\\
 National Institute of Technology Karnataka (NITK), Surathkal - 575025, India\\}}
\maketitle

\begin{abstract}
In today's day and age, we face a challenge in detecting deepfake images because of the fast evolution of modern generative models and the poor generalization capability of existing methods. In this paper, we use an ensemble of fine-tuned vision transformers like DINOv2, AIMv2 and OpenCLIP's ViT-L/14 to create generalizable method to detect deepfakes. We use the DF-Wild dataset released as part of the IEEE SP Cup 2025, because it uses a challenging and diverse set of manipulations and generation techniques. We started our experiments with CNN classifiers trained on spatial features. Experimental results show that our ensemble outperforms individual models and strong CNN baselines, achieving an AUC of 96.77\% and an Equal Error Rate (EER) of just 9\% on the DF-Wild test set, beating the state-of-the-art deepfake detection algorithm \textit{Effort} by 7.05\% and 8\% in AUC and EER respectively. This was the winning solution for SP Cup, presented at ICASSP 2025.
\end{abstract}


\section{Introduction}
Generative models today, have become an important part of a lot of applications across many industries. Synthetic data in many forms like audio, video and text has become widespread. Among the most concerning developments is the rise of deepfake facial imagery, which uses intricate generative algorithms to produce highly realistic but entirely fabricated human faces. These deepfakes pose a serious threat, as they can be used to spread misinformation, manipulate public perception, and damage trust in digital media. Popular generative models such as StyleGAN~\cite{NEURIPS2021_076ccd93}, ProGAN, and diffusion-based architectures like DALL·E and Stable Diffusion~\cite{rombach2022high} are frequently used to create such synthetic visuals. Modern image synthesis paradigms have become the foundation for a range of malicious applications, from generating fake faces entirely from scratch to enabling complex manipulations like face swapping and facial reenactment. 
These capabilities underscore the need for robust and reliable methods to detect and mitigate deepfake imagery. 

In this paper, we study two main features that could be used to detect deepfake images including pixels and features extracted using vision transformers (ViT). We then compare our approach to the existing state-of-the-art deepfake face detection algorithm~\cite{yan2024orthogonal} using the test set from the DF Wild Challenge, i.e., the IEEE Signal Processing Cup 2025.

\subsection{The Problem}
The key to building a practical solution for deepfake image detection is generalization. Generalization, here, refers to two aspects of deepfake detection; a model that is agnostic to the type of generator and a model that can detect purely synthetic images while also addressing manipulations like face-swapping and facial reenactment. Furthermore, techniques such as JPEG compression and image resizing often hamper the performance of existing deepfake detection algorithms.
\begin{figure*}
    \centering
    \includegraphics[scale=0.46]{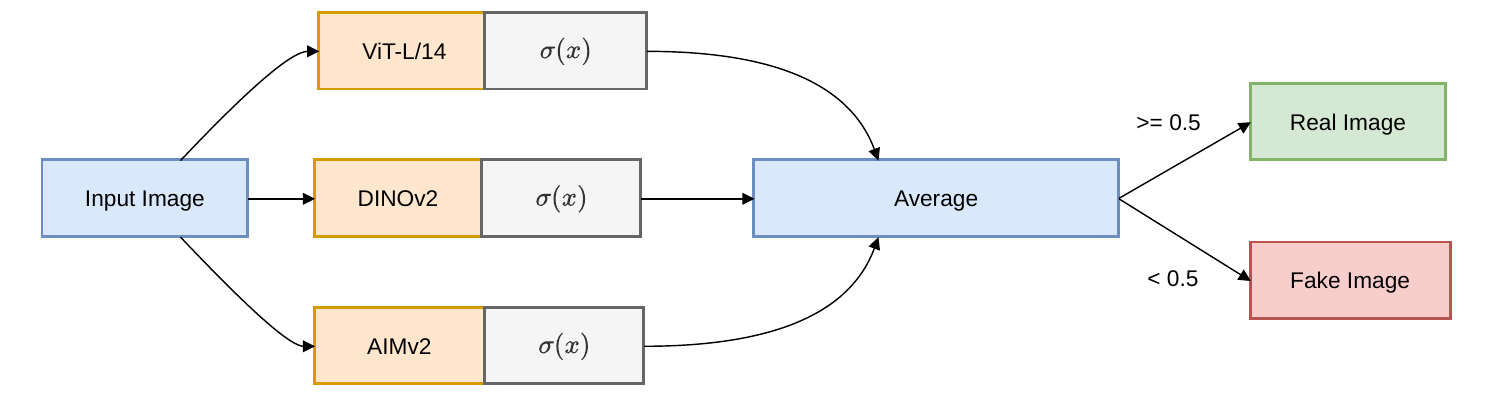}
    \caption{Our proposed ensemble of fine-tuned Vision Transformers for deepfake image detection}
    \label{fig:ensemble}
\end{figure*}
\subsection{Related Works}
In this section we hope to discuss existing approaches to synthetic image detection that have shaped the development of our methodology. While there are recent studies that have tried to tackle specific problems in deepfake image detection, our focus was on building a robust classifier for general-purpose synthetic image detection that would generalize and apply to Out-of-Distribution (OoD) datasets.

A prominent line of research for deep fake detection involves exploring frequency domain \cite{zhang2019detecting} cues as an effective means to distinguish real and synthetic images. Initial studies showed that deepfakes, particularly those generated by GANs, often display spectral anomalies, most likely as a product of the upsampling process that the images go through. As a result, frequency-aware detectors were proposed to capture these cues, often outperforming spatial-domain baselines.

Studies on frequency artifacts for deepfake detection~\cite{durall2020watch}~\cite{dzanic2020fourier} showed that these artifacts are easier to spot in the frequency domain. 
This line of research led to a better understanding of how frequency artifacts work and influenced not only detection techniques but also how image generators are designed. 
Wang~\cite{wang2020cnn2} proposed a detector based on the ResNet50 model, aimed at identifying fake images made using GANs. While this method worked well, it was mainly focused on GAN-generated images and did not perform as well on images from other generation methods like Diffusion Models. Frank~\cite{frank2020leveraging} built on this by using the Discrete Cosine Transform (DCT) to highlight frequency artifacts in fake images. Corvi et al.~\cite{corvi2023detection} built on this by using strong data augmentation to make their detector more robust and general. 
They trained their model on images from latent Diffusion Models and achieved strong results on both GAN and Diffusion Model-generated images.

\section{Methodology}
To address this issue, we used a variety of strategies, first concentrating on spatial features and then moving on to Vision Transformers like DINOv2~\cite{oquab2023dinov2}, AIMv2~\cite{fini2024multimodal}, and OpenCLIP's ViT-L/14~\cite{cherti2023reproducible}.

\subsection{Spatial Features}
A straightforward and effective method we employed involved training convolutional neural networks (CNNs) directly on raw pixel data from the DF Wild dataset. We experimented with multiple architectures, such as ResNet50 and DenseNet121.
Among these, ResNet50 was good at identifying subtle pixel-based artifacts, such as irregular textures, that are often found in synthetically generated images. The depth of the model allowed it to extract patterns that distinguish authentic content from fakes, leading to the classification performance shown in Table \ref{tab:model_comp}.

DenseNet121 offers a different architectural advantage due to its densely connected structure, where each layer integrates inputs from all previous layers. This allows for better feature reuse. This dense architecture enables the model to learn a wide variety of features at different levels of abstraction. Table \ref{tab:model_comp} shows the effectiveness of this approach.
\subsection{Vision Transformers}
To ensure the best performance of our classifier in this downstream task, we used vision foundation models such as DINOv2 from Meta AI and AIMv2 from Apple. Both DINOv2 and AIMv2 are self-supervised vision transformers that provide reliable representations for a range of computer vision applications after being pre-trained on extensive datasets. These foundation models are excellent candidates for enhancing deepfake detection, especially in real-world circumstances, since they have shown exceptional success in learning general-purpose visual characteristics.

Our goal was to improve these models using the DFWild-Cup competition dataset. We added two dense layers on top of the final transformer layers to modify these models for the real vs. fake image classification challenge. To improve the performance of our system, we used the pretrained (ViT/L-14) model in addition to DINOv2 and AIMv2. We loaded the pre-trained ViT/L-14 model, trained on extensive image datasets, using the same fine-tuning process as DINOv2 and AIMv2. These models were finetuned using an Adam optimiser with a learning rate of $10^{-6}$.

After fine-tuning DINOv2, AIMv2, and OpenCLIP ViT/L-14 individually, we took the next step to further boost our model's performance by creating an ensemble of all three models. Ensembles are known to improve robustness and accuracy by using the strengths of multiple models, each contributing unique features from the dataset. To form the ensemble, we averaged the final sigmoid outputs of each model discussed above. Figure \ref{fig:ensemble} shows a representation of the ensemble. This method helped reduce model-specific biases and variance, leading to a more balanced and accurate prediction. By allowing each model to contribute to the final classification, the averaging strategy improved the system's capacity to generalize unseen deepfake patterns. The ensemble approach proved to be the most effective method, giving the best performance in our experiments. This strategy outperformed any single model in isolation, ensuring that our solution could handle the varied and complex nature of the DFWild-Cup challenge dataset. The results of these approaches are presented in Table \ref{tab:model_comp}.

\section{Experimental Setup and Results}
All experiments for our deepfake detection system were conducted on a machine equipped with an Nvidia Quadro P5000 GPU (16GB VRAM), which, while not close to state-of-the-art by current standards, provided sufficient resources for training and evaluating our models with careful batch size and memory management. We employed Hugging Face Transformers to access and fine-tune pre-trained models like DINOv2, AIMv2, and ViT-L/14, leveraging their optimized implementations and pretrained weights. Scikit-learn was used for obtaining metrics. A weighted binary cross-entropy loss was used along with the Adam optimizer. 

\subsection{Dataset}
\begin{figure*}[h]
    \centering
    \begin{subfigure}{0.2\textwidth}
        \includegraphics[width=\linewidth]{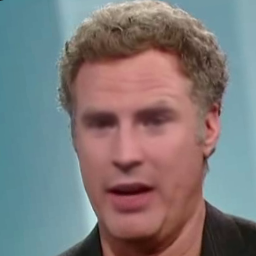}
    \end{subfigure}
    \begin{subfigure}{0.2\textwidth}
        \includegraphics[width=\linewidth]{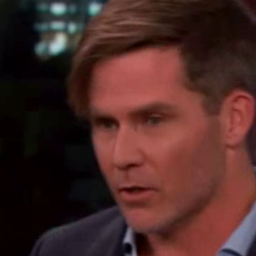}
    \end{subfigure}
    \begin{subfigure}{0.2\textwidth}
        \includegraphics[width=\linewidth]{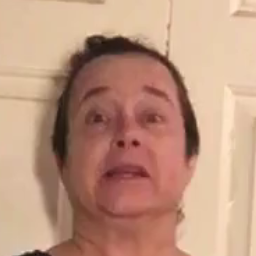}
    \end{subfigure}
    \begin{subfigure}{0.2\textwidth}
        \includegraphics[width=\linewidth]{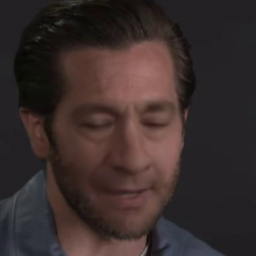}
    \end{subfigure}
    \begin{subfigure}{0.2\textwidth}
        \includegraphics[width=\linewidth]{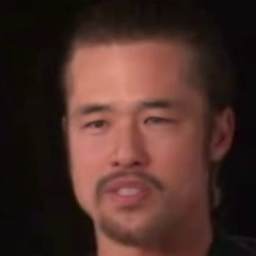}
    \end{subfigure}
    \begin{subfigure}{0.2\textwidth}
        \includegraphics[width=\linewidth]{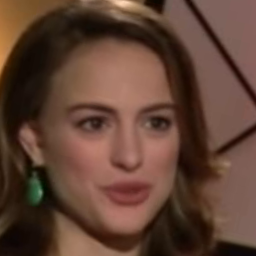}
    \end{subfigure}
    \begin{subfigure}{0.2\textwidth}
        \includegraphics[width=\linewidth]{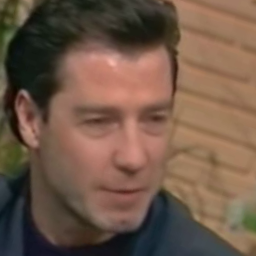}
    \end{subfigure}
    \begin{subfigure}{0.2\textwidth}
        \includegraphics[width=\linewidth]{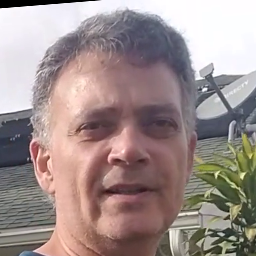}
    \end{subfigure}
    \caption{Images from the DF-Wild Challenge test set}
    \label{fig:five_images}
\end{figure*}
The DFWild challenge at ICASSP 2025 provided a comprehensive collection of deepfake datasets composed of many publicly available data sources. This compilation includes eight widely-used benchmark datasets: Celeb-DF-v1 and v2~\cite{Li_2020_CVPR}, FaceForensics++~\cite{roessler2019faceforensicspp}, DeepfakeDetection, FaceShifter~\cite{li2020advancing}, UADFV~\cite{li2018ictu}, Deepfake Detection Challenge Preview, and Deepfake Detection Challenge~\cite{dolhansky2020deepfake}. All datasets were pre-processed using a uniform pipeline to ensure that detectors do not implicitly learn dataset-specific or pre-processing-specific cues, thereby encouraging true generalization. To the best of our knowledge, this dataset hasn't been studied in existing literature. The training set consists of 42,690 real and 219,470 fake images, while the validation set includes 1,548 real and 1,524 fake images. The test set contains a total of 395,609 images with 76,594 real images and 319,015 fake images. Few sample images form the dataset are shown in Figure \ref{fig:five_images}

\begin{table}[h]
\centering
\caption{Model-level comparison of classifiers trained for the `Real vs. Fake' task. Parameters in Millions.}
\label{tab:cnn_inference_comparison}
\setlength{\tabcolsep}{8pt}
\begin{tabular}{|p{2.7cm}|c|p{1.3cm}|p{1cm}|}
  \hline
  \textbf{Model} & \textbf{\#Params} & \textbf{Inf. Time (ms)} & \textbf{Size (MB)} \\
  \hline
  \textit{DenseNet121} & 8.0  & 9.2  & 33 \\
  \hline
  \textit{ResNet50}    & 25.6 & 10.1 & 98 \\
  \hline
  \textit{DINOv2}      & 85   & 32.0 & 325 \\
  \hline
  \textit{AIMv2}       & 88   & 28.5 & 335 \\
  \hline
  \textit{AIMv2 + DINOv2} & 173 & 58.2 & 660 \\
  \hline
  \textit{AIMv2 + DINOv2 + ViT-L/14} & 418 & 124.5 & 1650 \\
  \hline
  \textit{Effort~\cite{yan2024orthogonal}} & 210 & 112.3 & 820 \\
  \hline
\end{tabular}
\end{table}

\begin{table*}[h]
\centering
\setlength{\tabcolsep}{8pt}
\begin{tabular}{|l|c|c|c|c|c|}
  \hline
  \textbf{Model} & \multicolumn{2}{c|}{\textbf{F1-score}} & \textbf{Accuracy} & \textbf{EER} & \textbf{AUC} \\
   & \textbf{Real} & \textbf{Fake} &  &  &  \\
  \hline
  \textit{DenseNet121} & 52.1 & 78.4 & 69.9 & 33.3 & 75.1 \\
  \hline
  \textit{ResNet50}    & 49.3 & 76.2 & 67.0 & 36.3 & 73.3 \\
  \hline
  \textit{ViT-L/14}    & 63.7 & 84.1 & 77.8 & 13.1 & 94.1 \\
  \hline
  \textit{DINOv2}      & 61.5 & 83.3 & 76.2 & 15.4 & 92.8 \\
  \hline
  \textit{AIMv2}       & 64.9 & 86.2 & 80.2 & 13.5 & 94.1 \\
  \hline
  \textit{AIMv2 + DINOv2} & 65.4 & 86.4 & 80.1 & 12.1 & 95.2 \\
  \hline
  \textit{\textbf{AIMv2 + DINOv2 + ViT-L/14}} & 71.2 & 89.3 & 84.4 & \textbf{9.0} & \textbf{96.8} \\
  \hline
  \textit{Effort} (Yan et. al.)~\cite{yan2024orthogonal} & 63.0 & 90.1 & \textbf{84.6} & 17.1 & 89.7 \\
  \hline
\end{tabular}
\caption{Comparison of classifiers trained for the `Real vs. Fake' task. All values are in \%.}
\label{tab:model_comp}
\end{table*}

\subsection{Results and Error Analysis}

We evaluated multiple classifiers for the Real vs. Fake detection task, with a focus on both traditional CNN backbones and transformer-based architectures. Table~\ref{tab:cnn_inference_comparison} summarizes model-level resource and latency characteristics for classifiers on the Real versus Fake task. The results reveal a clear trade-off between model complexity and inference cost: lightweight CNNs such as DenseNet121 and ResNet50 exhibit low memory footprints and fast inference times but have smaller parameter counts, whereas vision transformer models DINOv2, AIMv2 and ViT-L/14 require substantially larger parameter counts, storage and per-sample latency. Ensemble configurations further increase resource demands, with AIMv2+DINOv2 roughly doubling parameters and latency relative to single ViT models, and the full AIMv2+DINOv2+ViT-L/14 ensemble attaining the largest parameter count and model size. The Effort SOTA model occupies an intermediate position in this complexity spectrum. 

As seen in Table~\ref{tab:model_comp}, the results exhibit a clear performance disparity between the simpler convolutional baselines and more advanced transformer-based models, with the transformers performing much better.

Among the two CNN-based models, DenseNet121 performed best with an accuracy of 69.94\%. ResNet50, significantly underperformed on the test set with an accuracy of 67.05\% and an F1-score of just 49\% for Real and 76\% for Fake, indicating a strong bias toward the dominant (fake) class. This is likely due to the class imbalance in the training data, where fake images outnumber real ones by more than 5:1.

Transformer-based models demonstrated improved robustness and generalization, especially after fine-tuning them with the training set. DINOv2 achieved a notable boost over ResNet50, especially in AUC (92.84\%) and EER (15\%), though it still exhibited a low F1 for real images, suggesting persistent bias. Fine-tuned AIMv2 achieved the best standalone performance with F1-scores of 65\% for real and 86\% for fake images, along with an AUC of 94.06\%, indicating its strong generalization and insensitivity to class imbalance. The performance of fine-tuned OpenCLIP ViT-L/14 was marginally worse than that of AIMv2.

The naive ensemble of AIMv2 + DINOv2 achieved an F1-score for real images of around 65\% and an F1 for fake detection close to 86\%. Adding ViT-L/14 to the ensemble recovered fake detection performance and achieved the best EER (9\%) and AUC (96.77\%), with an accuracy of 84.41\%. Our proposed ensemble of these vision transformers outperformed \textit{Effort}~\cite{yan2024orthogonal} in both AUC and EER significantly while only showing marginally lower accuracy in comparison.

From these results, two primary trends emerge: Class imbalance has a tangible impact on model bias. Simpler models like ResNet50 heavily favor the fake class, while better-regularized or pretrained models such as AIMv2 and DINOv2 exhibit more balanced behavior. Model ensembling without careful calibration can harm class-specific performance. In particular, the AIMv2 + DINOv2 model’s failure on the fake class suggests representation-level conflicts or poor threshold alignment between models.

\section{Conclusion}
In this paper, we attempt to explore the problem of deepfake detection using a multitude of approaches from frequency features and spatial level-pixel representations to richer features extracted from VLMs. While we began with CNN architectures like ResNet and DenseNet, our focus shifted since our results demonstrated that while such methods perform reasonably well, their generalization capabilities are often limited. 

To address this, we moved to pretrained ViTs such as DINOv2, AIMv2, and ViT-L/14, using their feature representations. These models, when fine-tuned on our deepfake detection task,outperformed traditional approaches. Moreover, an ensemble of these models provided further improvement, underlining the value of combining complementary strengths for greater robustness and generalization.

\section{Future Work}

Future work will focus on including multimodal and continual-learning approaches to maintain robustness as generative models evolve. In particular, combining ViT-based image representations with audio, temporal, and metadata cues may help the system adapt more effectively to rapidly changing generative techniques.
One way to achieve this is by employing domain-adaptive calibration and uncertainty-aware online fine-tuning to accommodate novel generator families and common preprocessing shifts, such as compression and resizing. Parallel efforts can address practical deployment and interpretability by developing compact, quantized, or distilled ViT variants for low-latency edge inference, by producing token and region level attribution mechanisms and provenance-aware fingerprinting for forensic analysis, and by expanding evaluation protocols with adversarial, in-the-wild, and privacy preserving samples to more rigorously measure and improve out of distribution generalization.

\bibliographystyle{splncs04}
\bibliography{ref}

\end{document}